\UseRawInputEncoding
\documentclass[
]{ceurart}

\usepackage[utf8]{inputenc}
\usepackage{listings}
\usepackage{algorithm,algorithmic}
\usepackage{graphicx}
\usepackage{subcaption}
\usepackage[export]{adjustbox}
\usepackage{hyperref}

\begin{document}

\copyrightyear{2025}
\copyrightclause{Copyright for this paper by its authors.
  Use permitted under Creative Commons License Attribution 4.0
  International (CC BY 4.0).}

\conference{CLEF 2025 Working Notes, 9 -- 12 September 2025, Madrid, Spain}

\title{Improving Fungi Prototype Representations for Few-Shot Classification}

\tnotemark[1]

\author[1]{Abdarahmane Traore}[%
orcid=0000-0002-8942-4349,
email=eat4651@umoncton.ca,
url=https://abtraore.com
]
\cormark[1]

\author[1]{{\'E}ric Hervet}[%
email=eric.hervet@umoncton.ca,
]
\fnmark[1]

\author[1]{Andy Couturier}[%
email=andy.couturier@umoncton.ca,
]
\fnmark[1]

\address[1]{Embia, Computer Science Department, Faculty of Science, Université de Moncton, Moncton, N.B., Canada}

\cortext[1]{Corresponding author.}
\fntext[1]{These authors contributed equally.}

\begin{abstract}
The FungiCLEF 2025 competition addresses the challenge of automatic fungal species recognition using realistic, field-collected observational data. Accurate identification tools support both mycologists and citizen scientists, greatly enhancing large-scale biodiversity monitoring. Effective recognition systems in this context must handle highly imbalanced class distributions and provide reliable performance even when very few training samples are available for many species, especially rare and under-documented taxa that are often missing from standard training sets. According to competition organizers, about 20\% of all verified fungi observations, representing nearly 20,000 instances, are associated with these rarely recorded species. To tackle this challenge, we propose a robust deep learning method based on prototypical networks, which enhances prototype representations for few-shot fungal classification. Our prototypical network approach exceeds the competition baseline by more than 30 percentage points in Recall@5 on both the public (PB) and private (PR) leaderboards. This demonstrates strong potential for accurately identifying both common and rare fungal species, supporting the main objectives of FungiCLEF 2025.
\end{abstract}


\maketitle

\section{Introduction}

Accurate and efficient identification of fungal species is essential for scientific, ecological, and public health applications such as biodiversity monitoring and ecological research. Traditionally, this process has depended on the expertise of mycologists performing detailed morphological analyses. Such methods are often time-consuming and require specialized knowledge. The high diversity of fungal species and their subtle morphological differences further complicate manual identification.

The rise of citizen science platforms has led to a significant increase in the amount of observational data available for scientific analysis. This development offers a valuable opportunity to create automated recognition systems that can assist both specialists and amateurs. The FungiCLEF challenge, which is part of the LifeCLEF initiative ~\cite{Picek2024LifeCLEF, Picek2024Overview, lifeclef2025}, was established to address the technical challenges involved in automatic fungal species identification using real-world data.

Building on previous editions ~\cite{Picek2024FungiCLEFOverview, Picek2023FungiCLEFOverview, Picek2022FungiCLEFOverview}, the FungiCLEF 2025~\cite{fungiclef2025} competition focuses on some of the most difficult problems in the field. These include extremely imbalanced class distributions and the presence of many species with only a few available training samples. This situation creates a classic few-shot learning scenario. The challenge is particularly acute for rare and under-recorded species, which constitute a significant portion of fungal observations but are often poorly represented in curated datasets. According to the competition organizers, approximately 20\% of verified observations, or nearly 20,000 instances, correspond to these rarely recorded species. This statistic highlights the need for models capable of learning effectively from very limited data. Source code is available at  \href{https://github.com/abtraore/Prototype-Network-FungiCLEF25}{https://github.com/abtraore/Prototype-Network-FungiCLEF25}. 

To address these challenges, we present a deep learning approach that improves prototype representations for few-shot fungal species classification. Our method leverages advanced techniques to generate more discriminative and representative prototypes for each species, even when only a small number of training samples are available. We investigate the effectiveness of various pretrained models and few-shot learning frameworks on this task.

\begin{figure}[ht]
\centering
\includegraphics[width=15.5cm]{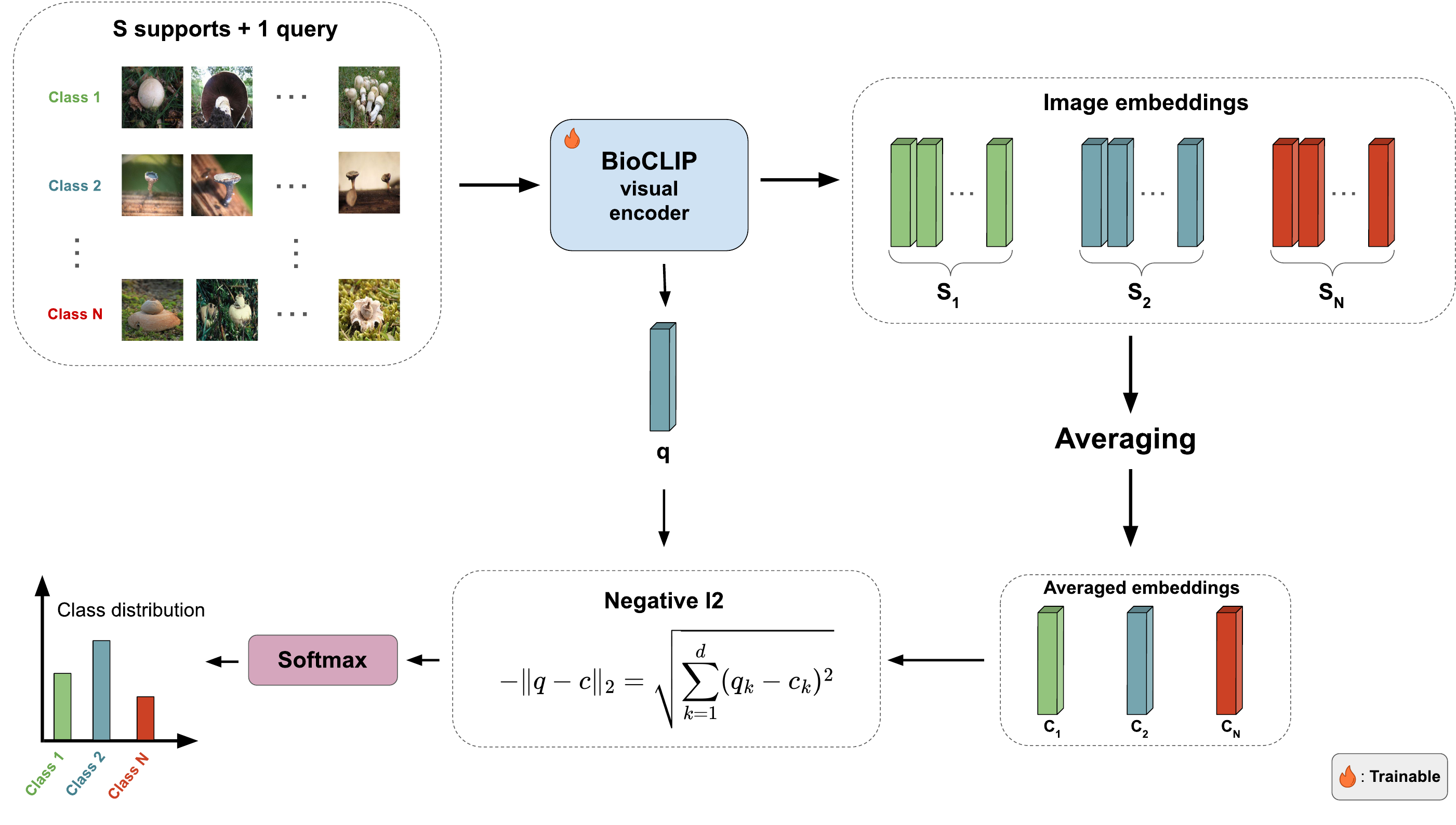}
\caption{Overview of the proposed architecture pipeline. For each class, a set of $S$ support images and one query image $q$ are encoded using the BioCLIP visual encoder to obtain image embeddings. Embeddings from the support set are averaged to form class prototypes. The distance between the query embedding and each class prototype is computed using the negative L2 metric, followed by a softmax operation to obtain class probabilities for the query.}
\label{fig:protonet}
\end{figure}

The primary contribution of this work is the development and the comprehensive evaluation of a method that provides a significant boost in few-shot classification accuracy on the FungiCLEF 2025 dataset. By refining prototype generation and implementing effective few-shot learning strategies, our approach achieves an improvement exceeding 30 percentage points over the competition baseline in Recall@5 on both the public and private leaderboards. These results demonstrate the strong potential of our approach for accurate identification of both common and rare fungal species. The outcomes directly support large-scale biodiversity monitoring efforts, particularly those involving citizen science initiatives. Our best results are obtained using a prototypical network approach, as illustrated in Figure~\ref{fig:protonet}.

\section{Related work}

Automated classification of fungi species using computer vision and machine learning has become a rapidly growing research area, fueled by the demand for quick and accessible identification tools for mycologists, citizen scientists, and ecological applications. Traditional methods are based on morphological examination, a process that is both time-consuming and dependent on significant expertise. Computational approaches seek to address these constraints by harnessing image data along with available metadata.

Earlier attempts in this field often used classical machine learning with hand-crafted features.
The introduction of deep learning, especially Convolutional Neural Networks (CNNs), has driven major advancements.
Architectures like ResNet, DenseNet, Inception, and EfficientNet have been widely used for image-based classification of fungi, showing high accuracy on targeted datasets~\cite{Ozsari2024, Arik2023}.
Some studies apply CNNs to classify pathogenic fungi from microscopic images~\cite{Kocagul2023}, while others use transfer learning and mobile platforms to identify mushroom species from field photographs~\cite{Arik2023}. Gaikwad et al. (2021) further demonstrate the use of CNNs to identify plant diseases caused by fungi via leaf images~\cite{Gaikwad2021}.

The FungiCLEF challenge, part of the LifeCLEF initiative, has played a major role in advancing research by providing large, well-annotated datasets and a standardized evaluation protocol~\cite{Picek2024LifeCLEF, Picek2024Overview}. These resources, often based on collections such as the Atlas of Danish Fungi, provide not only photographs but also rich metadata, supporting the development of multimodal classification systems.
The FungiTastic dataset~\cite{Picek2024}, for example, is a comprehensive multi-modal benchmark designed for closed-set, open-set, and few-shot fungi classification, highlighting the field’s complexity and fine granularity.

A central challenge in fungi classification lies in high variability within species and subtle differences between species, which creates a demanding fine-grained classification problem.
This difficulty is intensified by a long-tailed species distribution, where many taxa are represented by only a few observations in the available data.
Such challenges make few-shot learning approaches especially relevant.
While few-shot learning is widely studied in general image classification, its application to the nuanced and diverse domain of fungi has only recently begun to emerge.
Initial work has addressed plant diseases caused by fungi using few-shot techniques~\cite{Ouhami2022, Alfarrarjeh2024}, and more recently, some studies focus on rare fungi image classification directly~\cite{Su2024}.

Our research addresses the specific problem of few-shot fungal species classification in the context of the FungiCLEF 2025 challenge.
We build on the foundation of established models such as CLIP~\cite{Radford2021} and implement few-shot learning frameworks like Prototypical Networks~\cite{Snell2017}, with a focus on enhancing prototype-based representations.
By integrating pretrained vision models and methods suited to learning from extremely limited data, our goal is to accurately recognize both widespread and rare fungi species from real-world observations, further advancing the state-of-the-art in this challenging field.

\section{FungiCLEF 2025 challenge}

The FungiCLEF Challenge is designed to address the problem of few-shot recognition for fungi species using comprehensive real-world observational data. Each entry in the dataset includes several photographs of a single specimen, along with detailed metadata such as location, time, substrate, habitat, and toxicity.
The dataset also provides satellite images and weather-related variables.
The main goal is to develop a classification system that can generate a ranked list of species predictions for each observation, even when faced with the significant challenge of distinguishing among a large and diverse set of species. Many of these species are rarely observed or have very few available training examples. 

Participants are provided with a set of fungal observations to analyze, and their task is to return a list of the top predicted species from a given group of classes.
Automated identification of fungi has important implications for mycologists, nature enthusiasts, and citizen scientists by enabling species recognition in the field and supporting large-scale biodiversity cataloging efforts.
For these systems to be effective in widely used citizen science projects, they must be able to make accurate predictions with limited data and across a broad range of species, some of which are represented by only a handful of records. Additionally, rare and under-documented species are often missing from training datasets, increasing the challenge for artificial intelligence methods.
Analysis indicates that roughly 20 percent of all confirmed observations, or about 20,000 instances, are associated with such rare or poorly recorded species. This highlights the importance of developing models that can accurately identify them.

\subsection{Dataset}

The training and validation data for this challenge were sourced from fungi observations submitted to the Atlas of Danish Fungi up until the end of 2023. Each observation was reviewed and labeled by expert mycologists. Observations typically include several photographs of a single specimen, accompanied by detailed metadata such as timestamps, GPS coordinates, descriptions of substrate and habitat, and toxicity information. Furthermore, the dataset provides additional context through satellite imagery and meteorological data, creating a rich, multimodal resource for research and model development. Most observations are comprehensively annotated, ensuring that the dataset is both diverse and detailed. A complete description of the FungiTastic dataset is available in Picek et al.~\cite{Picek2024}.

\begin{figure}[htbp]
  \centering
  \begin{tabular}{cccc}
    \scriptsize Agaricus arvensis & \scriptsize Crocicreas furvum & \scriptsize Geastrum elegans & \scriptsize Menispora Caesia \\
    \includegraphics[width=.19\linewidth,height=.19\linewidth,keepaspectratio=false,valign=m]{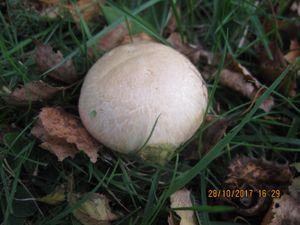} &
    \includegraphics[width=.19\linewidth,height=.19\linewidth,keepaspectratio=false,valign=m]{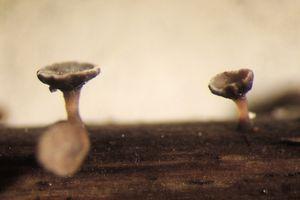} &
    \includegraphics[width=.19\linewidth,height=.19\linewidth,keepaspectratio=false,valign=m]{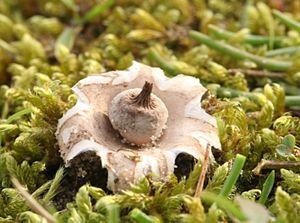} &
    \includegraphics[width=.19\linewidth,height=.19\linewidth,keepaspectratio=false,valign=m]{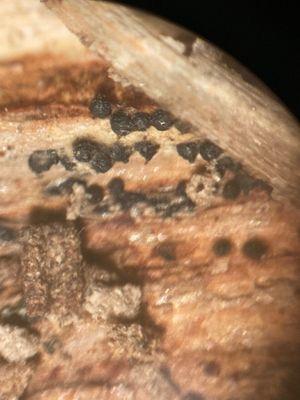} \\
    \rule{0pt}{2ex} & \rule{0pt}{2ex} & \rule{0pt}{2ex} & \rule{0pt}{2ex} \\
    \includegraphics[width=.19\linewidth,height=.19\linewidth,keepaspectratio=false,valign=m]{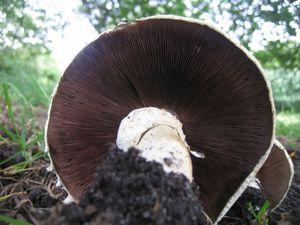} &
    \includegraphics[width=.19\linewidth,height=.19\linewidth,keepaspectratio=false,valign=m]{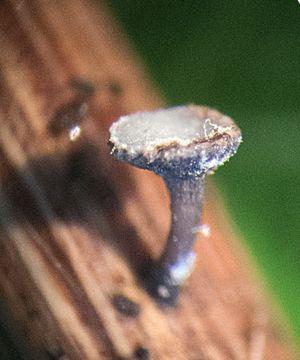} &
    \includegraphics[width=.19\linewidth,height=.19\linewidth,keepaspectratio=false,valign=m]{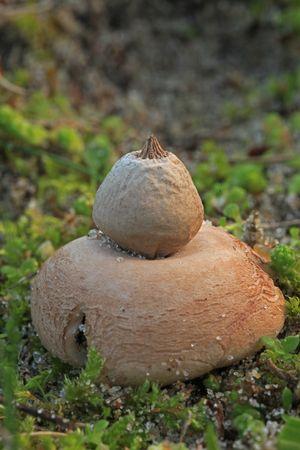} &
    \includegraphics[width=.19\linewidth,height=.19\linewidth,keepaspectratio=false,valign=m]{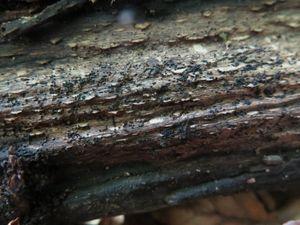} \\
    \rule{0pt}{2ex} & \rule{0pt}{2ex} & \rule{0pt}{2ex} & \rule{0pt}{2ex} \\
    \includegraphics[width=.19\linewidth,height=.19\linewidth,keepaspectratio=false,valign=m]{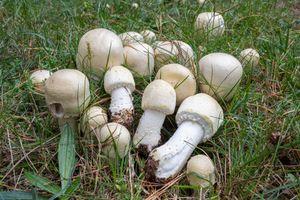} &
    \includegraphics[width=.19\linewidth,height=.19\linewidth,keepaspectratio=false,valign=m]{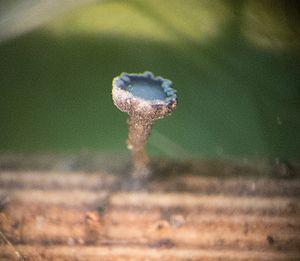} &
    \includegraphics[width=.19\linewidth,height=.19\linewidth,keepaspectratio=false,valign=m]{media/fungus/Geastrumelegans/0-2238464025.JPG} &
    \includegraphics[width=.19\linewidth,height=.19\linewidth,keepaspectratio=false,valign=m]{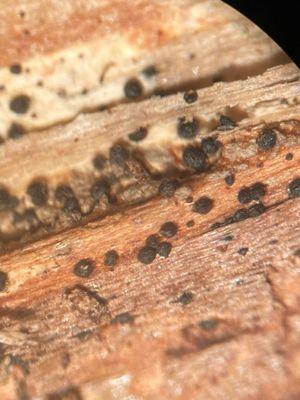} \\
  \end{tabular}
  \caption{Samples of Four Fungi Classes: Three representative images from each of four fungi classes (Agaricus arvensis, Crocicreas furvum, Geastrum elegans, and Menispora caesia), with one class per column.}
  \label{fig:dataset}
\end{figure}

For the challenge, participants received a well-structured dataset organized into several main folders. The "images" directory contains photographs at multiple resolutions (300px, 500px, 700px, and the original size); Figure~\ref{fig:dataset} shows a selection of sample images. The "metadata" directory provides separate files for the training, validation, and testing splits, with extensive attribute information for each observation. The "captions" directory includes machine-generated textual descriptions (Molmo-7B captions) for each image, stored in JSON format. It is important to note that these captions were made available only after the competition began. As a result, participants could utilize both structured metadata and, later, automatically generated captions, enabling multimodal approaches in their systems. The dataset is highly imbalanced, with each species represented by just one to four observations in the training set. Specifically, the training partition contains 4,293 observations and 7,819 images covering 2,427 unique species, while the validation set consists of 1,099 observations and 2,285 images across 570 classes.

\subsection{Evaluation Protocol}

The primary evaluation metric for this competition is Recall@\emph{k}, which measures the proportion of samples where the true species label appears among the top \(k\) predicted labels for each observation. It is formally defined as:

\[
\mathrm{Recall@}k = \frac{1}{N} \sum_{i=1}^{N} \mathbb{I}\left[ y_i \in \hat{Y}_i^k \right]
\]

where:
\begin{itemize}
  \item \(N\) is the total number of samples,
  \item \(y_i\) denotes the true label for the \(i\)-th sample,
  \item \(\hat{Y}_i^k\) is the set of top \(k\) predicted labels for the \(i\)-th sample,
  \item \(\mathbb{I}[\cdot]\) denotes the indicator function, returning 1 if the condition inside is true and 0 otherwise.
\end{itemize}

For the main evaluation, sets \(k=5\). Additionally, Recall@10 (\(k=10\)) will be reported for further analysis.

\section{Approaches}

To tackle the challenge of few-shot fungal species recognition, we evaluated a range of modeling strategies, progressing from simple baselines to more advanced methods. Our objective was to assess both established and novel approaches for maximizing classification performance given the constraint of very limited samples per species. We began by implementing prototype-based methods with pretrained models, establishing a straightforward yet informative point of comparison. Building on this baseline, we fine-tuned state-of-the-art CLIP architectures as introduced by Radford et al.~\cite{Radford2021}, leveraging both generated and refined captions alongside multimodal data to enhance representation learning and classification accuracy. Additionally, we explored prototypical networks following the methodology of Snell et al.~\cite{Snell2017}, which are well-suited for few-shot learning and enabled more effective recognition of rare and underrepresented species. The following sections detail our methodological framework, the models and techniques evaluated, and the results achieved with each approach.
As a reference, the competition organizers provided two non-parametric embedding-based baselines for few-shot classification using BioCLIP, a CLIP model fine-tuned on biological data. In both baselines, L2-normalized BioCLIP embeddings are computed for all train and test images. The first method ("BioCLIP + FAISS + NN") utilizes a nearest-neighbor classifier, where all training embeddings are indexed using FAISS, and each test image is assigned the label of its single closest training image in embedding space based on cosine similarity. The second method ("BioCLIP + FAISS + Prototypes") employs a prototype-based classifier, computing for each class a prototype vector defined as the mean (average) of all BioCLIP embeddings from the training images of that class. During inference, test images are assigned the label of the most similar class prototype according to cosine similarity. Both baselines require no additional training beyond feature extraction and serve to establish a reference performance for the challenge.
Notably, the competition organizers restricted their baselines to using only the train split as support images. For our experiments, we initially followed this protocol, but subsequently augmented our support set by incorporating images from the validation split alongside the training data, thereby providing additional examples per class and potentially improving classification robustness.

\subsection{Prototype Averaging (Baseline)}

As an initial benchmark, we implemented a prototype averaging approach using a selection of state-of-the-art pretrained models. Specifically, we employed OpenCLIP~\cite{Cherti2022}, which provides a comprehensive framework for feature extraction with several vision transformers, including both domain-specific and general-purpose variants. Among these, we evaluated BioCLIP~\cite{Stevens2023} (ViT-B-16-224, pretrained on TreeOfLife-10M), ViT-H-14-378-quickgelu (pretrained on DFN-5B), and ViT-B-16-224-quickgelu (pretrained on DFN-2B).

For this baseline, we extracted feature embeddings for each available image. For each class, we computed the prototype as the mean vector of the sampled embeddings. No further model training or finetuning was performed at this stage. Classification of test observations was based on the cosine similarity between the query embedding and all class prototypes.

We evaluated two strategies for prototype construction: using only training data, and using both training and validation data. Including the validation set for prototype averaging resulted in noticeably stronger performance, particularly for classes with few training examples. Table~\ref{tab:btable} summarizes the Recall@5 for both the public leaderboard (PB) and private leaderboard (PR), allowing for a thorough comparison of each method’s generalization ability.

The results demonstrate several important trends. First, augmenting prototypes with both training and validation data led to improved Recall@5 on both PB and PR. This is especially beneficial for classes with insufficient samples in the training set, as additional validation data helps provide more representative prototypes and reduces sensitivity to noise. Second, larger model architectures such as ViT-H-14-378-quickgelu consistently outperformed smaller ones. This highlights the impact of model capacity in the context of few-shot fungal classification. Notably, domain-specific pretraining with BioCLIP delivered competitive results, and on this task achieved the best performance, obtaining a Recall@5 of $0.6017$ on PB and $0.5472$ on PR. Finally, when using feature index-based retrieval methods such as FAISS~\cite{douze2024faiss}, we found that these approaches lagged behind prototype averaging for both PB and PR metrics in this few-shot scenario.

By providing Recall@5 on both public and private leaderboards, we ensure that our baseline evaluation reflects not only initial leaderboard feedback but also true generalization performance as measured on the hidden test split. This comprehensive evaluation, outlined in Table~\ref{tab:btable}, sets a robust foundation for subsequent improvements and comparisons.

\begin{table}
\caption{Recall@5 scores for prototype averaging and feature-based retrieval methods using different pretrained models on the few-shot fungi classification task. The table compares performance on the public (PB) and private (PR) leaderboards, demonstrating the effects of model architecture and dataset pretraining.}
  \label{tab:btable}
  \begin{tabular}{ccccc}
    \toprule
    Models/Methods  & Pretrained Dataset & \text{Recall@}5 (PB) & \text{Recall@}5 (PR)\\
    \midrule
    BioCLIP (ViT-B-16-224) & TreeOfLife-10M dataset & \textbf{0.6017} & \textbf{0.5472} \\
    ViT-H-14-378-quickgelu & DFN-5B & 0.5221 & 0.4928 \\
    ViT-B-16-224-quickgelu & DFN-2B  & 0.46902 & 0.44631\\
    BioCLIP + FAISS + Prototypes & TreeOfLife-10M dataset & 0.33185 & 0.26649 \\
    BioCLIP + FAISS + NN & TreeOfLife-10M dataset & 0.28318 & 0.24708 \\
  \bottomrule
\end{tabular}
\end{table}

\subsection{CLIP Finetuning}

To generate image captions for our dataset, our initial approach involved incorporating metadata into prompts and explicitly requesting models to describe visual appearance. In these early experiments, we used Molmo-7B~\cite{Deitke2024}. However, analysis of the results showed that Molmo-7B often hallucinated visual characteristics and sometimes fabricated metadata extracted from prompts. As a result, we discontinued its use for this task. To reduce such inconsistencies, we instead selected GPT-4.1-2025-04-14~\cite{OpenAI2023GPT4} via the OpenAI API. This model allowed us to systematically identify and filter out incoherent or inaccurate description content, producing more precise and detailed image captions. When using GPT-4.1, we observed that many generated captions were too long for direct use, often exceeding the typical context window for CLIP’s text encoder, which is generally truncated to approximately 77 input tokens. To ensure compatibility for CLIP finetuning, we instructed GPT-4.1 to summarize and shorten the captions as needed.

To determine whether high-quality captions could also be generated efficiently using consumer hardware, we performed supervised finetuning (SFT) with the HuggingFace SFT library~\cite{wolf-etal-2020-transformers} on the lightweight Qwen2-VL-2B~\cite{Wang2024} model. We used LoRA quantization~\cite{hu2022lora} with rank 8, $\alpha=16$, and a dropout rate of $5 \times 10^{-2}$. The Qwen2-VL-2B model was trained using five epochs, a per-device batch size of four, gradient accumulation steps of eight, a learning rate of $2 \times 10^{-4}$, a maximum gradient norm of $3 \times 10^{-1}$, a warmup ratio of $3 \times 10^{-2}$, cosine learning rate scheduling, and bf16 mixed precision. For all CLIP finetuning experiments, we utilized the ViT-H-14 model pretrained on DFN-5B, with a warmup period, batch size of five, learning rate of $1 \times 10^{-5}$, weight decay of $0.01$, ten training epochs, automatic mixed precision, and cosine learning rate scheduling.

All experiments employed the fullsize images subset of our dataset. Images were resized as necessary to match network input dimensions using bicubic interpolation, which preserved image quality and minimized artifacts. We evaluated the performance of the models finetuned on captions from both GPT-4.1 and Qwen2-VL-2B on the test set. Table~\ref{tab:ctable} presents the Recall@5 scores for both the public leaderboard (PB) and the private leaderboard (PR).

The results show that finetuning on GPT-4.1-generated captions led to the highest retrieval effectiveness, achieving a Recall@5 of $0.54424$ on PB and $0.48382$ on PR. In comparison, the model trained on Qwen2-VL-2B captions reached a Recall@5 of $0.41150$ on PB and $0.36093$ on PR. Both methods outperform the FAISS-based retrieval baselines, but the gap between finetuning with GPT-4.1 captions and Qwen2-VL-2B captions is substantial on both leaderboards. The difference in performance between the public and private leaderboards is expected and reflects the challenge of generalizing to the hidden (private) test split. Notably, the relative ranking of the models is maintained from PB to PR, confirming that improvements from high-quality caption supervision are robust and not an artifact of leaderboard overfitting. This comparison also demonstrates that the quality and richness of supervision strongly affect retrieval performance and model generalization.

Overall, these findings confirm that generating high-quality, carefully constructed captions is critically important for cross-modal retrieval performance on both the public and hidden benchmarks. At the same time, results also show that advances in resource-efficient models like Qwen2-VL-2B are promising, though high-capacity models such as GPT-4.1 remain the most effective as supervision sources for few-shot fungi classification.

\begin{table}
\caption{
Recall@5 results for few-shot fungi classification using CLIP models finetuned with various captioning strategies and FAISS-based baselines. Performance is shown for both the public (PB) and private (PR) leaderboards to highlight the effect of caption source on model accuracy.
}
  \label{tab:ctable}
  \begin{tabular}{ccccc}
    \toprule
   Models/Methods  & Finetuned Caption & \text{Recall@}5 (PB) & \text{Recall@}5 (PR)\\
    \midrule
    ViT-H-14-378-quickgelu  & Qwen2-VL-2B  & 0.41150 & 0.36093 \\
    ViT-H-14-378-quickgelu  & GPT-4.1-2025-04-14 & \textbf{0.54424} & \textbf{0.48382} \\
    BioCLIP + FAISS + Prototypes & TreeOfLife-10M dataset & 0.33185 & 0.26649 \\
    BioCLIP + FAISS + NN  & TreeOfLife-10M dataset & 0.28318 & 0.24708 \\
  \bottomrule
\end{tabular}
\end{table}

\subsection{Prototypical Networks}

Prototypical networks are highly effective for few-shot classification tasks. Their core principle is to learn an embedding space in which samples from the same class cluster around a central point, termed the \emph{prototype}. Each episode during training consists of sampling $K$ classes (ways), with $S$ support samples (shots) and $Q$ query samples per class. All samples are processed by a neural network to generate embeddings. The prototype for each class, $\mathbf{c}_k$, is computed as the mean of the support embeddings:
\[
    \mathbf{c}_k = \frac{1}{S} \sum_{i=1}^{S} f_\theta(\mathbf{x}^{(k)}_i),
\]
where $f_\theta$ denotes the embedding function and $\mathbf{x}^{(k)}_i$ are the support samples of class $k$. Each query sample is then embedded and classified by identifying the closest prototype based on a distance metric, such as squared Euclidean distance:
\[
    d(\mathbf{x}_q, \mathbf{c}_k) = \| f_\theta(\mathbf{x}_q) - \mathbf{c}_k \|^2.
\]
The network is trained to minimize a loss that encourages query embeddings to be closest to their corresponding class prototypes, facilitating generalization to new classes from few examples.

During inference, all images from the training and/or validation splits are passed through the prototypical network to compute embeddings for every observation in each class. For evaluation, test samples are also embedded and compared to the stored class prototypes. Here we use cosine similarity to measure closeness. The top-5 nearest classes are then reported for each query, supporting Recall@5 evaluation.

For our experiments, we systematically varied the key parameters: number of classes per episode ($K$), number of support samples per class ($S$), and the number of query samples per episode ($Q$). Following best practices, we matched these values between training and evaluation phases, as consistency is known to yield better generalization. Previous work also suggests that increasing $K$ can improve performance, given sufficient computational resources.

In our first experiment, we trained BioCLIP with $K=70$ classes and $S=3$ support samples per class over $500$ episodes using the training split. This setup achieved a Recall@5 of $0.61061$ on the public leaderboard and $0.57179$ on the private leaderboard, outperforming all prior methods.

We further evaluated the ViT-B-16-SigLIP-512 model, which accepts $512 \times 512$ input images compared to BioCLIP's $224 \times 224$. Due to GPU memory constraints, we reduced $K$ to $15$ for these experiments and trained for $750$ episodes. This configuration yielded Recall@5 scores of $0.53539$ (PB) and $0.53298$ (PR), lower than BioCLIP but still better than non-episodic baselines.

Our best results were achieved by optimizing our code to support $K=75$ for BioCLIP, training for $750$ episodes with $S=3$ and $Q=1$. This approach yielded the highest Recall@5 values: $0.64159$ on the public leaderboard and $0.58473$ on the private leaderboard. The value of $K$ is primarily constrained by available GPU memory, and further increases may deliver additional gains.

All prototypical network experiments utilized Stochastic Weight Averaging (SWA)~\cite{Izmailov2018}, starting parameter accumulation $100$ episodes before training ended, with an initial learning rate of $1 \times 10^{-5}$ and an SWA learning rate of $5 \times 10^{-5}$.

As summarized in Table~\ref{tab:ptable}, prototypical networks consistently outperformed all previous baselines and finetuned CLIP models on both public and private leaderboards. While performance on the private leaderboard was slightly lower, the key trends and rankings were stable. The moderate reduction in Recall@5 from public to private splits suggests that these models generalize well to unseen data. Overall, our findings highlight episodic training with prototypical networks as a highly effective strategy for the challenging, large-class, and imbalanced few-shot fungi classification task.

\begin{table}
\caption{
Recall@5 performance of prototypical network models for few-shot fungi classification, showing the effects of varying episode parameters (K, S, Q), encoder type, and training split. Results for both public (PB) and private (PR) leaderboards are presented, along with FAISS-based baselines for comparison.
}
  \label{tab:ptable}
  \begin{tabular}{ccccccccc}
    \toprule
   Models/Methods & K & S & Q   & Training split & \text{Recall@}5 (PB) & \text{Recall@}5 (PR) \\
    \midrule
        BioCLIP (ViT-B-16-224)  & 75 & 3 & 1  & train/val & \textbf{0.64159} & \textbf{0.58473} \\
        BioCLIP (ViT-B-16-224)  & 70 & 3 & 1  & train & 0.61061 & 0.57179 \\
        ViT-B-16-SigLIP-512  & 15 & 3 & 1   & train & 0.53539 & 0.53298 \\
        BioCLIP + FAISS + Proto  & NA & NA & NA & train & 0.33185  & 0.26649 \\
        BioCLIP + FAISS + NN  & NA & NA & NA & train & 0.28318 & 0.24708 \\
  \bottomrule
\end{tabular}
\end{table}

\section{Results Discussion}

\begin{figure}[ht]
  \centering
 \includegraphics[width=15.5cm]{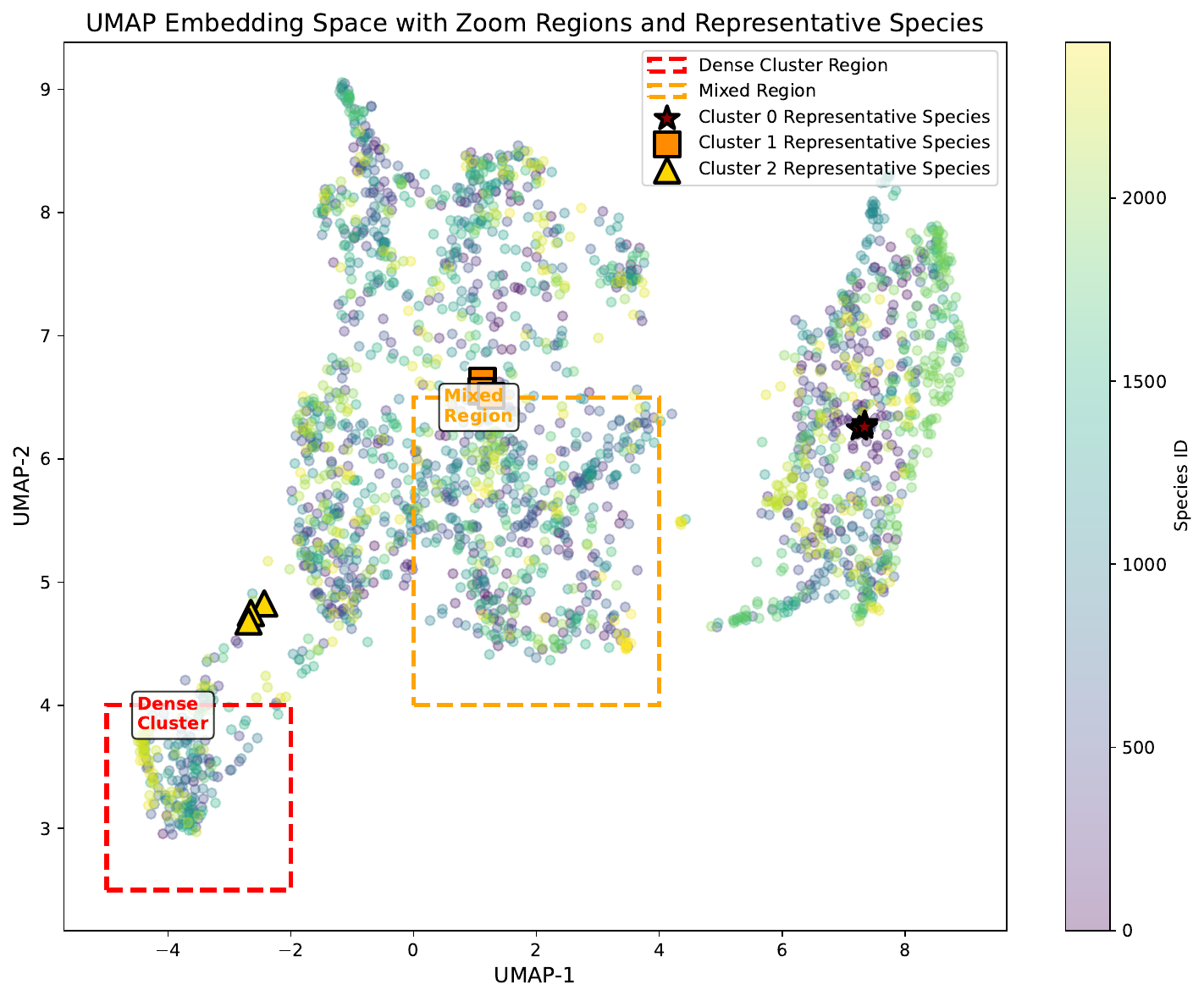}
\caption{UMAP embedding space of 2,427 fungal species prototypes from our best-performing model. The dense cluster region (red dashed rectangle) and mixed region (orange dashed rectangle) highlight areas of particular interest. Representative species from three automatically identified clusters (K-means) are marked with star, square, and triangle symbols. These regions demonstrate different embedding characteristics: the dense cluster shows tightly packed related species, while the mixed region represents the convergence of different taxonomic groups.}
\label{fig:umap-embeddings}
\end{figure}

An in-depth examination of our experiments reveals several important trends regarding the effectiveness and generalization of the methods we assessed. Throughout this analysis, we reference Recall@5 results for both the public leaderboard (PB) and the private leaderboard (PR) to enable clear, robust comparisons of performance.

The prototype averaging baseline shows that increasing the number of samples per class prototype by including both training and validation data leads to noticeable performance gains. This effect is especially pronounced for classes with very limited training observations, where extra samples help reduce noise and promote more reliable classification. For instance, using BioCLIP, the Recall@5 achieves $0.6017$ on PB and $0.5472$ on PR. These results establish a solid benchmark for few-shot classification in this domain. The difference between PB and PR is moderate. This suggests some expected drop in performance when moving from the publicly visible portion of the test set to the hidden set, but it also confirms the underlying reliability of the approach.

Further gains are obtained using CLIP finetuning, particularly when supervised with high-quality, visually accurate captions generated by advanced language models such as GPT-4.1. Here, models trained on GPT-4.1 captions attain Recall@5 of $0.5442$ on PB and $0.4838$ on PR. These results clearly outperform models trained with Qwen2-VL-2B-generated captions, as well as all FAISS-based retrieval baselines. Importantly, the absolute performance is consistently higher on PB than on PR. However, the ordering of methods remains unchanged, supporting the robustness of improvements achieved by better supervision and model selection, rather than by leaderboard-specific tuning.

The best performance is achieved with prototypical networks when trained in high-way, few-shot episodic settings. For example, BioCLIP configured with 75 classes per episode and three support samples per class attains Recall@5 values of $0.6416$ on PB and $0.5847$ on PR. These results mark a significant improvement over all baselines and finetuned models. They highlight the benefit of episodic training and prototype-based classification for generalizing to new classes with limited data. As before, there is a small reduction in recall from PB to PR, but the overall advantage remains clear across both leaderboards.

To better understand why prototypical networks achieve such strong performance, we visualized the learned embedding space using UMAP dimensionality reduction (Figure~\ref{fig:umap-embeddings}). The visualization reveals that our best-performing model successfully organizes the 2,427 fungal species into distinct, well-separated clusters. Three major groupings emerge in the embedding space, with smooth transitions between related species within each cluster. The dense cluster region shows particularly tight packing with gradual color transitions, indicating that the model has learned to group morphologically similar species. In contrast, the mixed region where different clusters meet contains species from diverse taxonomic groups that share certain visual features. This hierarchical organization explains our model's high recall performance: when classifying a query image, taxonomically related species with similar visual characteristics are naturally positioned nearby in the embedding space, increasing the likelihood that the correct species appears within the top-5 predictions.

Overall, the results demonstrate several central principles. Both the quantity and quality of supervision, whether through increased data diversity or advanced captioning, play a crucial role in recall under highly imbalanced, few-shot training conditions. Episodic training with prototypical networks proves especially effective for these large-scale, rare-class biodiversity tasks. The learned representations, as visualized through UMAP, confirm that the model captures meaningful biological relationships between species. Consistent trends across both PB and PR provide strong evidence of the genuine generalization achieved by our models, as opposed to overfitting on visible evaluation data. These findings define a strong benchmark for future research in automated fungal species recognition under realistic and data-limited circumstances.

\section{Conclusion}

In this work, we conducted a systematic evaluation of prototype averaging, CLIP finetuning with caption-based supervision, and episodic prototypical networks for few-shot fungal species classification as posed by the FungiCLEF 2025 challenge. Our experiments show that the combination of diverse, high-quality training data and advanced few-shot learning frameworks leads to substantial improvements over the competition baseline. The best results were obtained by prototypical networks trained with episodic sampling, which consistently outperformed both standard prototype averaging and finetuned CLIP approaches.
We achieved improvements of more than 30 percentage points in Recall@5 on both the public and private leaderboards. This demonstrates that our approach is not only effective on the visible evaluation split, but also achieves robust generalization to the hidden test set. The consistent trends in performance across both leaderboards indicate that our methods address overfitting and ensure real-world applicability.
By providing detailed comparisons and reporting public as well as private leaderboard results throughout, we ensure that our findings are both transparent and reliable. This approach directly supports the goals of biodiversity monitoring and improving citizen science platforms, especially in settings with highly imbalanced data and very limited annotations for many species. Our results establish strong benchmarks and offer insights for future research into automated recognition under challenging, data-limited conditions.

\section{Declaration on Generative AI}

While preparing this work, the authors made use of GPT-4.1 to assist with paraphrasing, rewording, and performing grammar and spelling checks. The authors subsequently reviewed and revised the content as appropriate and accept full responsibility for the final published material.

\bibliography{sample-ceur}
\end{document}